# When the Target Domain Changes: AI-Mediated Construct Drift in High-Stakes English Language Assessment

Yi Gui
Measurement Incorporated
queenmg@gmail.com

## Abstract

High-stakes English proficiency tests treat standardized, unaided performance as evidence for score interpretations about academic English proficiency. This interpretation remains meaningful, but as target language use domains increasingly involve generative AI, the extrapolation from unaided test performance to academic communicative readiness becomes less self-evident. This conceptual validity argument reframes AI as a score-interpretation problem in high-stakes language testing, not only an operational issue of scoring, feedback, security, or misconduct. Synthesizing current literature in three uneven layers, the paper shows that most work treats AI as assessment infrastructure, while far less theorizes its implications for construct validity and extrapolation warrants. It defines AI-mediated construct drift as the misalignment that arises when communicative abilities required in the target domain change through AI mediation while test constructs remain anchored to an unaided-performance model. It proposes bounded AI mediation as a validity-oriented design principle: a standardized condition in which all test takers access the same institutionally controlled AI assistant, with predefined assistance boundaries, logged interactions, and tasks that distinguish comprehension support from answer generation. The paper argues that score interpretations should be narrowed and supplemented when used to support claims about AI-mediated academic communication.

**Keywords:** construct validity; English language assessment; artificial intelligence; large language models; target language use; extrapolation; authenticity; consequential validity; construct drift; bounded AI mediation

## Toward Bounded AI Mediation as a Validity-Oriented Design Principle

High-stakes English language proficiency tests have historically played a gatekeeping role in international higher education. Universities, professional programs, and immigration systems use TOEFL, IELTS, the Duolingo English Test, Pearson PTE, and related assessments to support decisions about whether test takers have sufficient English proficiency to study, work, or participate in English-medium institutions. These tests differ in format, scoring, delivery, and theoretical tradition, but they share a broad interpretive logic: performance under standardized test conditions is used to support a score interpretation about academic English proficiency, and that interpretation is then extrapolated to English-medium academic or professional communication.

The rapid spread of generative artificial intelligence (AI) and large language models (LLMs) has placed pressure on this interpretive logic. Much public and professional discussion has treated AI as a problem of cheating, plagiarism, automated scoring, test preparation, or test security. Those concerns are real. Test providers must attend to security, fairness, proctoring, data privacy, score comparability, and the reliability of AI-supported scoring systems. Yet for language testing, the sharper measurement problem is whether AI is changing the target language use (TLU) domain to which English proficiency scores are interpreted to generalize.

In academic settings, students increasingly read with AI assistance, draft and revise with feedback tools, summarize complex texts, translate across languages, prepare oral presentations with language support, and collaborate with AI systems during writing, research, and study. These practices are unevenly distributed, variably regulated, and still evolving. They are not uniformly legitimate in all institutions or assignments. Nevertheless, when such practices become part of the communicative ecology of academic work, the extrapolation warrant linking unaided standardized performance to academic communicative readiness becomes more conditional.

This is therefore a score-interpretation problem in high-stakes language testing, not simply an AI problem. Traditional English proficiency scores may continue to provide useful evidence of baseline linguistic resources. A student who cannot read, write, listen, or speak in English without support may face serious barriers in English-medium study even if AI tools are available. The claim advanced here is narrower: unaided English proficiency scores should not automatically be interpreted as full evidence of academic communicative readiness in environments where academic communication is substantially AI-mediated.

The paper makes three contributions. First, it specifies how AI-mediated domain change weakens specific inferences in the existing validity argument for high-stakes English proficiency scores. Second, it defines AI-mediated construct drift as a validity problem that arises when target-domain communicative demands change faster than test constructs and score interpretations. Third, it proposes bounded AI mediation as a validity-oriented design principle for future validation research. Bounded AI mediation is not a full test design and not a claim that AI-mediated testing is already valid. It is a conceptual middle path between total AI prohibition and unrestricted AI access.

The argument is conceptual rather than empirical. It does not present new data, conduct a meta-analysis, or offer a systematic review. Instead, it uses validity theory, technology-based assessment guidance, and an integrative synthesis of recent AI and language assessment literature to analyze where traditional score interpretations may need to be narrowed, clarified, or supplemented. A conceptual validity argument should not overstate what the literature can support. It should make explicit what is being challenged, what remains meaningful, and what future evidence is needed.

## The Traditional Construct: Unaided Academic English Proficiency

The traditional construct underlying high-stakes English proficiency testing can be described, in broad terms, as individual academic English proficiency demonstrated under standardized and largely unaided conditions. This is an intentionally simplified formulation. TOEFL, IELTS, DET, PTE, and other assessments operationalize the construct differently. Some emphasize integrated tasks, some use automated scoring, some include human interviewers or raters, and some are more adaptive or machine-scored than others. Still, for admissions and placement purposes, the core claim is similar: the test taker's performance under controlled conditions provides evidence about their ability to use English in academic settings.

This construct made sense for several reasons. First, academic participation has historically required substantial independent control of English. Students must read course materials, follow lectures, participate in discussion, write assignments, communicate with instructors, and complete assessments. Second, high-stakes decisions require comparability. Testing under standardized conditions reduces variation in access to help, time, tools, interlocutors, and institutional resources. Third, unaided performance provides evidence of baseline language resources. Even in tool-rich environments, students still need enough linguistic competence to judge whether a translation is plausible, whether an AI-generated summary is accurate, whether a revision changes meaning, and whether a spoken contribution is appropriate.

The problem, therefore, is not that unaided English proficiency was never meaningful. Nor is the problem that English proficiency no longer matters. Rather, the problem is that the extrapolation domain may be changing. A score that supports the claim "this test taker can perform certain academic English tasks under standardized unaided conditions" does not automatically support the broader claim "this test taker is ready for AI-mediated academic communication." The second claim may require additional evidence about tool-mediated reading, writing, revision, source evaluation, multilingual mediation, and ethical use of AI.

This distinction follows directly from contemporary validity theory, particularly the unified framework in which construct validity subsumes all forms of validity evidence (Messick, 1989). The Standards for Educational and Psychological Testing define validity in relation to evidence and theory supporting interpretations of test scores for proposed uses (American Educational Research Association et al., 2014). Validity is not a permanent property of a test form. It is tied to a score interpretation, a population, a context, and a use. If multiple interpretations are intended, each needs support. Technology-based assessment guidance makes a similar point: technology can improve assessment, but it can also introduce construct underrepresentation, construct-irrelevant variance, access barriers, and fairness concerns (International Test Commission & Association of Test Publishers, 2022). Bachman and Palmer (2010) formalized this logic in their Assessment Use Argument (AUA) framework, in which the validity of a language assessment is justified through a chain of claims linking test performance to domain description, score interpretation, decisions, and consequences. Their concept of the target language use (TLU) domain—the specific real-world settings requiring language

performance—is directly relevant: if the TLU domain changes, the foundational claim of the assessment argument is destabilized. Chapelle et al. (2008) applied a similar argument-based approach to the redesigned TOEFL, demonstrating how validity evidence can be organized around the interpretive claims a test score is meant to support.

The AI-era question is therefore not "Are TOEFL and IELTS valid?" in a global sense. The question is more precise: Which interpretations of high-stakes English proficiency scores remain well supported when academic communication is increasingly AI-mediated, and which interpretations require new evidence?

## Three Uneven Layers of AI and Language Assessment Research

The current literature can be organized into three uneven layers. This organization is more useful than an article-by-article summary because it highlights how the field's empirical energy is distributed.

### Layer 1: AI as Operational Infrastructure

The dominant layer treats AI as operational infrastructure for existing assessment systems. In this layer, AI is studied as a scorer, feedback provider, item generator, proctoring tool, security mechanism, quality-control system, or administrative support. The central question is whether AI can make existing assessment practices more efficient, reliable, scalable, secure, or fair.

Automated scoring and feedback studies illustrate the pattern. Recent work on ChatGPT and related LLMs for IELTS-style writing and L2 essays reports partial alignment with human or official ratings, alongside score discrepancies, lower or unstable ratings, limits in feedback specificity, and subgroup fairness concerns (Chen et al., 2025; Huang & Wilson, 2025; Koraishi, 2024; Saricaoglu & Bilki, 2025; Uyar & Buyukahiska, 2025; Wilson & Huang, 2024). The important validity point is not whether such systems can be useful; they may be useful for formative feedback, second-opinion scoring, and operational efficiency. The point is that agreement with existing ratings is not construct validation. If rubrics or criterion measures already underrepresent the target domain or embed subgroup bias, AI systems trained to approximate them may reproduce the inherited construct assumptions.

Similar cautions apply to automarking, remote proctoring, and responsible-AI governance. Work on these topics highlights efficiency, consistency, security, privacy, fairness, transparency, accountability, human oversight, and continuous monitoring (Galaczi & Pastorino-Campos, 2025; International Test Commission & Association of Test Publishers, 2022; Isbell et al., 2023; Johnson, 2025; Xu et al., 2024). This layer is necessary but insufficient for the present argument: it asks how AI can support, scale, or protect existing assessment paradigms, while usually leaving the target construct and extrapolation warrant intact.

### Layer 2: AI as an Authentic Academic Communication Ecology

A smaller but theoretically important layer treats AI as part of the communicative environment outside the test. Here, the question is not whether AI can score old tasks, but whether academic language use itself is changing. AI-assisted writing, reading, translation, summarization, feedback, prompt design, and human-AI collaboration are increasingly part of academic work. This does not mean they are always permitted or pedagogically desirable. It does mean that a serious target-language-use analysis can no longer assume that academic communication is always unaided.

Xi (2023) makes this issue explicit by arguing that AI use by test takers, not only AI use by test providers, pushes the field to reconsider constructs, tasks, rubrics, score interpretations, and uses. Xi identifies different stances toward technology in construct definition, ranging from treating technology as construct-irrelevant, to including limited digital skills, to redefining language constructs as communication integrated with digital and AI-mediated literacy. Xi (2025) extends this argument by proposing an AI-mediated interactionalist approach to communicative competence that includes AI digital literacy, effective AI tool use, and the interpretation and application of AI-generated outputs and feedback.

The present paper shares Xi's (2023, 2025) concern that AI requires rethinking language assessment constructs, but takes a different analytical route. Xi (2025) develops a construct-theoretical account of what communicative competence may need to encompass in AI-mediated environments, proposing an AI-mediated interactionalist approach that includes AI digital literacy, effective AI tool use, and the interpretation and application of AI-generated outputs and feedback. The present paper analyzes the validity consequences for existing high-stakes score interpretations when that domain change outpaces assessment revision. Grounded in Kane's (2013) argument-based validity framework, it asks not what should be added to communicative competence but where existing interpretive claims require renewed backing. The two approaches address different immediate evidential tasks. Xi's framework requires defining the expanded construct before it can be assessed; the argument-based approach pursued here requires that existing score claims be examined against a changing domain—enabling the field to identify which interpretations need renewed evidence even before the full contours of AI-mediated communicative competence are established. Furthermore, Xi (2025) suggests that high-stakes testing faces considerable challenges in adopting an expanded construct and recommends beginning with low-stakes formative assessments. The bounded AI mediation principle proposed here offers a middle path: it does not require full construct expansion but does require standardizing the mediation condition when the intended score interpretation refers to AI-mediated academic communication.

Perrotta et al. (2025) provide a related perspective from IELTS test preparation. They treat prompting not merely as a technical strategy but as a form of human-computer communication and emerging literacy. This matters for construct validity because prompt formulation, output evaluation, revision judgment, and rhetorical decision-making may become part of academic communication. Jin and Fan (2023) likewise argue that test taker engagement in AI technology-mediated language assessment should be considered in validation, because test takers' perceptions, strategies, and technology-mediated practices can affect both performance and consequences.

Chuang and Yan's (2025) systematic review of GenAI in language assessment also points to this shift. Their review finds growing work on scoring, feedback, validation, and fairness, but comparatively little on assessment planning and construct reconceptualization. They call attention to new tasks, criteria, authenticity, multimodal GenAI, ecological validity, and long-term impact. This is a crucial field-level pattern: the literature is moving faster on AI applications than on the construct theory needed to interpret them.

**Layer 3: AI as a Construct-Validity Challenge**

The third layer treats AI as a challenge to construct validity, score interpretation, and extrapolation. This layer remains underdeveloped, but it is the layer most relevant to the present paper.

Several sources show that the pre-AI extrapolation baseline was already imperfect. Isaacs et al. (2023) found that DET scores predicted some postgraduate academic outcomes in a major UK university context, but patterns were weaker or absent in some undergraduate groups and were context-dependent. Xie (2024) showed that integrated reading-to-write tasks broadened construct representation and added predictive value for academic writing, but also that prediction remained modest and complex. Barkaoui (2025) found that relations between IELTS or TOEFL scores and GPA varied by test and discipline over time in a large longitudinal sample. These studies do not show that English proficiency tests are invalid. They show that extrapolation from proficiency scores to academic outcomes is already partial, local, and criterion-sensitive.

AI intensifies that pre-existing problem. If academic communication increasingly involves AI-assisted reading, writing, summarizing, translation, and feedback, then the criterion domain changes. A score obtained under unaided conditions may still be useful, but it may no longer exhaust the evidence needed for claims about academic communicative readiness. This is the construct-validity challenge. The issue is not only whether AI introduces cheating or whether AI scoring is reliable. The issue is whether the target construct represented by the test is still aligned with the communicative ecology to which scores are extrapolated. From the perspective of Bachman and Palmer’s (2010) TLU domain analysis, the question becomes concrete: what language use tasks do students actually perform in current academic settings, and to what extent do those tasks now involve AI mediation? Without updated domain evidence, the gap between what tests assess and what academic communication requires cannot be empirically bounded.

Table 1 summarizes the three layers.

**Table 1**

*Three Uneven Layers of AI and Language Assessment Research*

| Layer | How AI is treated | Typical examples | Dominant validity question | Limitation for the present problem |
|---|---|---|---|---|
| Layer 1: AI as operational infrastructure | AI supports, scales, secures, or automates existing assessment systems | Automated scoring, feedback, item generation, proctoring, identity verification, responsible-AI governance | Can AI improve reliability, efficiency, security, fairness, or feedback quality within the existing paradigm? | Often preserves the pre-AI construct and inherited rubrics |
| Layer 2: AI as authentic communication ecology | AI is part of how students read, write, translate, revise, summarize, and collaborate | Prompting, AI-assisted writing, AI feedback, translation, multilingual mediation, human-AI collaboration | What does academic communication look like when AI tools are part of the environment? | Often not connected to high-stakes score interpretation |
| Layer 3: AI as construct-validity challenge | AI changes the domain to which test scores are extrapolated | Communicative competence in the AI era, construct representation, authenticity, extrapolation, consequences | Do unaided scores still support broad claims about academic communicative readiness? | Still underdeveloped in task models, rubrics, validation designs, and score reports |

## Argument-Based Validity Analysis: Where the Traditional Inference Weakens

Following Kane's (2013) argument-based approach, the traditional score interpretation can be represented as a chain of inferences:

Unaided standardized test performance -> academic English proficiency -> ability to communicate in English-medium academic settings -> appropriate admissions, placement, or support decisions.

This chain is not necessarily invalid. It is also not self-validating. Each link rests on warrants that require evidence. AI-mediated academic communication creates pressure at several points.

First, domain definition becomes less stable when AI mediates academic communication. Bachman and Palmer (2010) defined the target language use (TLU) domain as a specific setting outside the test that requires the test taker to perform language use tasks, and argued that test tasks should correspond as closely as possible to tasks in that domain. Their Assessment Use Argument framework makes domain description the foundation of the entire validity chain: if the domain is misdescribed, every subsequent inference is weakened. In traditional high-stakes testing, the TLU domain was characterized primarily as unaided academic English communication—reading scholarly texts, participating in seminars, writing essays, and delivering presentations without technological mediation. If the academic domain now includes AI-assisted reading, writing, translation, summarization, feedback, and knowledge navigation, then the traditional domain description becomes incomplete. Critically, domain change does not require that all academic communication becomes AI-mediated. It requires only that enough academic communication involves AI tools that the domain description underlying the test can no longer be assumed to represent the full range of relevant communicative practices.

Second, task representation becomes more contested. A tool-prohibited task provides clean evidence of baseline proficiency, but it may underrepresent authentic academic communication in contexts where AI use is expected or permitted. Chapelle (2001) identified authenticity as one of the central criteria for evaluating technology-mediated language assessment—the degree to which test task characteristics correspond to features of target language use. In the pre-AI era, the authenticity of standardized writing tasks was already debatable: students rarely write timed essays without access to sources, dictionaries, or revision time. AI tools intensify this authenticity gap. If a graduate student routinely drafts arguments with AI-assisted outlining, revises with AI feedback, and verifies claims with AI search tools, then a timed essay under proctored conditions may represent a narrower slice of the target domain than the score interpretation implies. Xie's (2024) findings are instructive: even before explicit AI access, integrated reading-to-write tasks broadened construct representation relative to independent writing. AI-mediated academic writing introduces further dimensions—source evaluation, prompt refinement, output verification—that unaided tasks may not elicit.

Third, evaluation and scoring become more complex in two distinct ways. When AI scoring systems are used to evaluate unaided performance, they may approximate human ratings without engaging construct validity. Agreement with human ratings is necessary but not sufficient; if the rubric itself underrepresents the target construct, both human and AI scorers will reproduce that limitation. Furthermore, if AI-mediated tasks are introduced, scoring must accommodate new dimensions of performance. A test taker who effectively refines a weak AI draft through targeted prompting, identifies a hallucinated claim, and reorganizes the argument around verified sources has demonstrated different abilities than one who accepts the first AI output uncritically. Huang and Wilson (2025) showed that prompt engineering can improve LLM-based scoring, but subgroup fairness remains a concern. Wilson and Huang (2024) found that automated scores for elementary English learners were not uniquely biased relative to human scores but did reproduce biases present in human rating. These findings suggest that the evaluation challenge is not simply "AI bias" but rather inherited limitations in scoring frameworks that were designed for unaided performance.

Fourth, explanation becomes less straightforward because performance on AI-mediated tasks cannot be attributed solely to language proficiency. In an AI-mediated task, performance may reflect language knowledge, prompt strategy, critical evaluation of AI output, tool familiarity, source-use judgment, and awareness of institutional AI policies. This multidimensionality creates an explanatory challenge: how much of the observed performance is attributable to the construct of interest? Chapelle et al. (2008) addressed an analogous challenge in building the validity argument for the redesigned TOEFL, where they needed to demonstrate that internet-based delivery did not introduce construct-irrelevant features. The AI-mediated case is more complex because the tool is not merely a delivery medium but an active interlocutor that shapes the communicative task itself. Response-process evidence—think-aloud protocols, interaction logs, keystroke data, revision traces—becomes essential for supporting explanatory claims about what the score actually measures.

Fifth, extrapolation becomes more conditional because the relationship between test performance and real-world academic communication may differ across AI-mediated and unaided conditions. This inference has always been the hardest link in the validity argument for high-stakes proficiency testing. AI mediation makes the warrant more context-dependent.

Sixth, utilization and consequences become more consequential because score users—admissions officers, program directors, immigration officials—may overinterpret baseline English proficiency scores as full evidence of academic readiness. When academic environments increasingly expect AI-mediated communication, a score report that states only "Writing: 25/30" provides no information about whether the test taker can function in those environments. Score reports may need to specify the conditions under which performance was elicited and the boundaries of the interpretive claim. This is consistent with Bachman and Palmer's (2010) emphasis on the consequences warrant in their Assessment Use Argument: the chain of reasoning must extend from score to decision to outcome, and each link requires evidence. If institutions use language test scores to predict success in AI-mediated academic programs, they need evidence that the scores support that particular prediction—not only the narrower claim about unaided proficiency.

Table 2 presents the analysis in argument-based form.

**Table 2**

*AI-Era Pressures on a Traditional Validity Argument*

| Validity inference | Traditional warrant | AI-era pressure | Needed evidence |
|---|---|---|---|
| Domain definition | English-medium academic communication requires sufficient individual academic English proficiency | Academic communication may increasingly include AI-assisted reading, writing, translation, summarization, feedback, and collaboration | AI-era target language use studies by discipline, institution, policy context, task type, and student population |
| Task representation | Standardized, unaided tasks sample relevant academic language abilities | Tool-prohibited tasks may represent baseline resources but underrepresent AI-mediated academic communication | Comparisons of unaided and AI-mediated tasks; content analyses of real course assignments and AI policies |
| Evaluation and scoring | Scores reflect the intended language construct and can be compared across test takers | AI scoring, proctoring, and tool interfaces may introduce construct-irrelevant variance or reproduce human bias | Human-machine agreement, subgroup fairness, response-process evidence, explainability, model-version documentation |
| Explanation | Performance can be explained primarily as academic English proficiency | AI-mediated performance may also reflect prompt strategy, output evaluation, source verification, tool familiarity, and ethical judgment | Think-aloud studies, interaction logs, keystroke data, revision traces, prompt-quality analyses |
| Extrapolation | Test performance supports inferences about communication in English-medium academic settings | AI-mediated academic communication may depend on strategies not elicited by unaided tests | Predictive studies using AI-supported academic communication outcomes, not GPA alone |
| Utilization and consequences | Cut scores and score reports support admissions, placement, and support decisions | Score users may overinterpret unaided scores as full academic readiness; AI policies may create inequitable consequences | Consequential-validity studies, institutional policy alignment, score-reporting experiments |

The table does not imply that the traditional inference collapses. It implies that the inference should be bounded. A traditional score can still support the interpretation that a test taker has demonstrated baseline academic English proficiency under specified conditions. It may not, without additional backing, support a broader claim about readiness for AI-mediated academic communication.

This distinction is especially important for high-stakes score users. Institutions often want a single number to support admissions or placement. But if the target domain is changing, a single score may need a clearer interpretive label. “Unaided academic English proficiency” is a narrower and more defensible claim than “academic communicative readiness” in a broad sense.

## AI-Mediated Construct Drift

AI-mediated construct drift is defined here as follows:

AI-mediated construct drift occurs when the communicative abilities required in the target language use domain change because of AI mediation, while test tasks, rubrics, and score interpretations remain anchored to an earlier unaided-performance model. In this paper, drift refers to a change in the relation between score interpretation and target-domain demands, not to longitudinal score instability or measurement drift in the psychometric sense (e.g., item-parameter drift or loss of measurement invariance).

The concept does not claim that English proficiency disappears, that TOEFL, IELTS, DET, or other tests become meaningless, that all AI-mediated practices are legitimate academic communication, or that every test must immediately allow AI. It names a narrower validity risk: the target domain may change faster than the assessment argument used to interpret scores.

Three features make construct drift useful as a validity concept: temporal divergence, psychometrically invisible misalignment, and systemic weakening of the validity argument.

First, construct drift is temporal. A test may have adequately represented its target domain when it was designed, but the domain shifts while the test, its rubrics, and its score interpretations remain anchored to an earlier model. The resulting misalignment is not simply an original design flaw; it is a consequence of domain change outpacing assessment revision.

Second, construct drift can remain invisible to standard psychometric monitoring. Score reliability can remain high, interrater agreement can be stable, and correlations with other language measures can persist while the score's extrapolation meaning becomes less secure. Traditional indicators may not flag the problem because they are calibrated within the assessment system or against criteria that share the same domain assumptions. Renewed TLU analysis is therefore needed to detect whether the domain itself has moved.

Third, construct drift is systemic. It does not weaken only one inference. When the domain description becomes outdated, task representation loses authenticity, evaluation criteria miss emerging dimensions of performance, explanatory claims become less complete, extrapolation evidence becomes less applicable, and score-use decisions rest on increasingly narrow evidence.

This concept overlaps with, but is not reducible to, established validity terms. Construct underrepresentation (Messick, 1989) refers to a test's failure to capture important aspects of the construct at a given moment; construct drift explains how such underrepresentation can emerge over time. Authenticity mismatch refers to a gap between test tasks and target language use features; construct drift names the case in which that gap results from domain change rather than a deliberate design tradeoff. Domain evolution is the descriptive fact that communicative practices change; construct drift is the validity consequence when assessment systems and score interpretations do not keep pace.

When construct drift occurs, its consequences appear across the validity argument. Traditional unaided tasks may underrepresent AI-mediated academic communication, unrestricted AI access may introduce construct-irrelevant variance through unequal tools or prompt literacy, total AI prohibition may deepen authenticity mismatch, and the same score may carry different meanings across institutions, disciplines, policy regimes, and time periods.

The practical implication is that high-stakes language testing may need a two-track validity strategy. One track preserves and clarifies unaided baseline proficiency. The other develops evidence for AI-mediated academic communication where such communication is part of the target domain.

## Bounded AI Mediation as a Validity-Oriented Design Principle

Bounded AI mediation is introduced here as a validation design principle, not an operational recommendation that high-stakes tests should immediately permit AI. It addresses a research problem created by construct drift: how can assessment researchers study AI-mediated academic communication without either prohibiting the tool conditions that define the domain or allowing uncontrolled access that makes scores difficult to interpret?

Bounded AI mediation is a standardized testing condition in which all test takers have access to the same institutionally controlled AI assistant, with predefined assistance boundaries, logged interactions, and task designs that distinguish support for comprehension, clarification, and revision from direct answer generation or authorship replacement.

The key idea is not to “let students use ChatGPT” in an unrestricted way. The key idea is to standardize the mediation condition itself so that AI-mediated academic communication becomes researchable. In AI-mediated academic environments, fairness may require not only prohibiting tools but also standardizing, bounding, and documenting the mediation conditions under which performance is elicited.

As a validation design principle, bounded AI mediation has several features.

First, tool access is standardized. All test takers interact with the same assistant, under the same interface, model version, system prompt, language settings, access permissions, and time limits. This reduces construct-irrelevant variance from unequal tool access.

Second, assistance boundaries are predefined. The assistant may be allowed to clarify vocabulary, explain instructions, summarize source passages at a specified level, flag coherence problems, or provide revision suggestions. It may be prohibited from generating full answers, writing thesis statements, inventing evidence, or producing final prose for submission. The boundary depends on the intended construct.

Third, interactions are logged. Logs can provide response-process evidence, support score interpretation, help detect misuse, and make prompt strategy visible. Logs should not be treated automatically as scores; they are evidence to be validated.

Fourth, tasks are designed to distinguish support from authorship replacement. For example, a reading task might allow bounded clarification of source text but require the test taker to evaluate claims. A writing task might permit feedback on clarity but require justification of accepted and rejected suggestions. A synthesis task might include AI-generated summaries with embedded inaccuracies and ask test takers to verify them against sources.

Fifth, scoring should separate product and process. A bounded AI-mediated task may assess the final communicative product, baseline linguistic control, source use, prompt strategy, evaluation of AI output, revision judgment, and ethical disclosure. Whether all of these should be scored in a high-stakes English proficiency test is an open question. The present paper does not argue that TOEFL or IELTS should absorb the entire construct. It argues that these dimensions need validation if score users want to make broader claims about AI-mediated academic communication.

Table 3 contrasts three AI-access conditions in terms of validity advantage, validity risk, and score interpretation.

**Table 3**

*Three AI-Access Conditions and Their Validity Tradeoffs*

| Design option | Validity advantage | Validity risk | Best-supported interpretation |
|---|---|---|---|
| Total AI prohibition | Strong comparability; clear evidence of unaided baseline proficiency; easier security model | May underrepresent AI-mediated academic communication; may create authenticity mismatch | Test taker can perform specified academic English tasks without AI assistance |
| Unrestricted AI access | High ecological similarity to some real-world tool-rich environments | Unequal access, uncontrolled model differences, prompt literacy variance, authorship ambiguity, weak score comparability | Difficult to interpret without extensive process evidence |
| Bounded AI mediation | Standardizes the mediation condition; allows targeted evidence about AI-mediated communication; creates process logs | Not yet validated; boundaries may be artificial; scoring may become complex; risk of construct creep | Test taker can perform specified academic communication tasks under controlled AI mediation |

To illustrate what bounded AI mediation might look like in practice, consider one prototype: an AI-assisted source-based writing task. Test takers receive two academic source texts on a specified topic and access to a standardized AI assistant embedded in the test platform. The assistant can clarify vocabulary on request, summarize individual paragraphs of the source texts at a controlled level of detail, and flag coherence problems in the test taker's draft. It cannot generate thesis statements, produce full paragraphs of argument, fabricate evidence, or write concluding sections. One source text contains an AI-generated summary with two embedded factual inaccuracies relative to the original. Test takers must identify the inaccuracies, use the sources to correct them, and write a short argumentative synthesis integrating evidence from both sources.

This task design would allow evidence about several construct-relevant dimensions: reading comprehension of academic sources, evaluation of AI-generated content against primary texts, strategic use of AI clarification to support rather than replace understanding, and written argumentation under process constraints. Scoring could separate the final written product (assessed for argumentation quality, source integration, and linguistic control), hallucination detection accuracy, and process evidence from interaction logs (number and type of AI queries, acceptance or rejection of suggestions). The key validity claim would be narrowly specified: this score represents the test taker's ability to produce a source-based argumentative synthesis in English using bounded AI assistance, not a general measure of AI readiness or academic communication in an unrestricted sense.

This prototype is illustrative, not prescriptive. Detailed task designs, scoring rubrics, generalizability studies, and operational feasibility analyses remain for future work. The prototype is meant to make the validity problem researchable, not to prescribe immediate operational adoption.

## Native/Non-Native Asymmetry and the Risk of Construct Expansion

Any proposal to include AI-mediated academic communication in English proficiency assessment must confront a fairness problem: native/non-native asymmetry. High-stakes English proficiency tests are usually required of international or non-English-background students, not of native English speakers entering the same academic environments. If AI-mediated academic communication is folded only into tests for non-native speakers, testing programs may create construct creep. International students would be required to demonstrate not only unaided baseline proficiency but also broader AI-mediated academic readiness, while native English-speaking students may face no comparable gatekeeping assessment.

This risk matters because AI-mediated academic readiness is not only an English-language issue. Prompting, source verification, hallucination detection, ethical disclosure, and human-AI collaboration may be relevant to all students. Some dimensions are language-mediated, but others are general academic skills. Those broader skills should not be silently absorbed into English proficiency testing unless the language-specific portion of the construct is clearly justified.

A more defensible approach is to distinguish three constructs:

1. Unaided English baseline proficiency. This refers to the ability to perform academic English tasks without AI assistance. It remains meaningful and may remain essential for admissions and support decisions.

2. AI-mediated English academic communication. This refers to using English with bounded AI support to read, write, revise, summarize, evaluate, and communicate in academic contexts.

3. General AI-mediated academic readiness. This includes broader tool literacy, ethical judgment, source verification, and academic integrity practices that may apply to all students, native and non-native alike.

This distinction helps prevent overexpansion of English proficiency tests. It also clarifies score reporting. A test score should not silently move from the first construct to the second or third. If an assessment claims to measure AI-mediated academic communication, it must specify whether the construct is language-specific, academic-generic, or both.

## Research Agenda

This paper's proposal is intentionally conceptual. Bounded AI mediation is a validation design principle, not a validated test. AI-mediated construct drift is a hypothesis about score interpretation under domain change, not a completed empirical finding. The next step is a validation research agenda.

First, the field needs AI-era target language use domain analyses. Researchers should examine how students and faculty actually use AI for reading, writing, translation, summarization, discussion preparation, feedback, and revision across disciplines and institutions. These analyses should attend to institutional policies, assignment types, student language backgrounds, and faculty expectations. Without domain evidence, construct redesign will remain speculative.

Second, studies should compare unaided and AI-mediated task conditions. The same learners could complete parallel tasks under no-AI, bounded-AI, and broader-AI conditions. Researchers could examine score structure, reliability, response processes, subgroup differences, task difficulty, and relations to external criteria. If AI-mediated performance has a different dimensional structure or predicts different AI-supported academic communication outcomes, that would support the construct-drift argument.

Third, response-process studies should use AI interaction logs. Logs can show what test takers ask, what suggestions they accept or reject, how they revise, and whether they detect errors. However, logs create privacy and fairness concerns. Their use must be governed by clear consent, security, and scoring principles.

Fourth, scoring research should examine prompt quality, hallucination detection, revision judgment, and source verification. These dimensions are not yet established high-stakes language assessment constructs. They require rubrics, rater training, generalizability evidence, fairness studies, and external validation.

Fifth, subgroup fairness must be studied under standardized AI assistance. Researchers should examine whether bounded AI mediation narrows or widens gaps by language background, socioeconomic status, disability, AI literacy, prior schooling, and institutional context. Wilson and Huang (2024) show why this matters: automated scoring may replicate human or institutional bias rather than introduce a clearly separate AI-specific bias.

Sixth, predictive validity studies should use AI-supported academic communication outcomes. GPA alone is too broad and noisy. Better criteria might include source-based writing assignments under known AI policies, instructor ratings of communicative readiness, revision tasks, oral presentations, AI-use process logs, disciplinary writing outcomes, and longitudinal academic support needs.

Seventh, score reporting models should distinguish unaided proficiency from AI-mediated academic communication. Reports might state the assistance condition under which performance was elicited, the intended interpretation, the limits of extrapolation, and the contexts for which evidence is strongest.

Table 4 summarizes this agenda.

**Table 4**

*Future Validation Research for AI-Mediated Construct Drift*

| Research priority | Core question | Possible evidence |
|---|---|---|
| AI-era TLU analysis | What does academic communication look like under current AI policies and practices? | Faculty interviews, student logs, assignment analyses, policy reviews, disciplinary case studies |
| Unaided vs. AI-mediated task comparison | Does AI access change score structure, difficulty, reliability, or criterion relations? | Experimental task-condition studies, generalizability analysis, many-facet models, multi-group analyses |
| Response-process evidence | How do test takers use AI during performance? | Prompt logs, revision traces, think-alouds, screen recordings, interaction coding |
| Scoring new dimensions | Can prompt refinement, output evaluation, and hallucination detection be scored reliably and validly? | Rubric studies, rater training, interrater reliability, construct modeling, criterion studies |
| Subgroup fairness | Does bounded AI assistance reduce or amplify inequities? | DIF, predictive-bias analyses, subgroup score comparability, access-condition studies |
| Predictive validity | Do scores predict AI-supported academic communication outcomes? | Longitudinal studies using course writing, presentations, source-based assignments, support needs |
| Score reporting | How should score users interpret baseline and AI-mediated evidence? | Score-report experiments, stakeholder interviews, decision studies |
| Policy alignment | How do institutional AI policies affect score use? | Cross-institutional validation, policy mapping, consequential-validity studies |

## Conclusion

High-stakes English proficiency tests should not be discarded simply because AI has entered academic life. Unaided academic English proficiency remains meaningful, and in many settings it remains necessary. Students still need linguistic resources to interpret, evaluate, revise, and communicate, even when AI tools are available. The stronger argument is not obsolescence, but interpretive narrowing.

As academic communication becomes increasingly AI-mediated, the traditional inference from unaided test performance to broad academic communicative readiness becomes less self-evident. Validity theory requires score interpretations to be tied to intended uses and target domains. If the target domain changes, the score interpretation needs renewed evidence. This is the central problem named here as AI-mediated construct drift.

The field has made substantial progress in studying AI as operational infrastructure. It has begun to study AI as part of communication. It has only begun to study AI as a construct-validity challenge for high-stakes score interpretation. Bounded AI mediation offers one way to move the conversation forward: not by abandoning unaided proficiency, not by allowing unrestricted AI access, and not by prescribing immediate operational adoption, but by standardizing, bounding, and documenting mediation conditions when the intended construct requires evidence about AI-mediated academic communication.

The next validity question is therefore not whether AI should be banned or embraced in a general sense. It is what claims a score is meant to support, what communicative domain those claims refer to, and what evidence is needed when that domain changes.